\documentclass[10pt,letterpaper]{article}
\usepackage{graphicx}
\usepackage{multirow}
\usepackage{booktabs}
\usepackage{placeins} 
\usepackage[table]{xcolor}
\usepackage{cogsci}
\usepackage{xurl}
\cogscifinalcopy 

\usepackage[
  style=apa,
  natbib=true,
  annotation=false,
]{biblatex}
\usepackage{float} 

\usepackage{ifthen}
\usepackage{calc}
\usepackage{siunitx}
\makeatletter
\newcommand{\absnum}[1]{%
  \begingroup
    \dimen0=\dimexpr#1pt\relax
    \ifdim\dimen0<0pt
      \dimen0=-\dimen0
    \fi
    \edef\abs@temp{\strip@pt\dimen0}%
    \num{\abs@temp}%
  \endgroup
}
\newcommand{\updiff}[1]{%
  \ifthenelse{\NOT\lengthtest{#1 pt < 0 pt}}{%
    \textsuperscript{%
      \colorbox{green!10}{\scriptsize$\uparrow$\,\absnum{#1}}%
    }%
  }{%
    \textsuperscript{%
      \colorbox{red!10}{\scriptsize$\downarrow$\,\absnum{#1}}%
    }%
  }%
}
\makeatother

\newcommand{\headcell}[2]{
  \parbox[c][1.88em][c]{\widthof{\textbf{#1}}}{
    \centering \textbf{#1}\\[-0.1em]\small #2
  }
}

\renewcommand{\paragraph}[1]{\vspace{.1em}\noindent\textbf{#1}}

\title{DELTA: Deliberative Multi-Agent Reasoning with Reinforcement Learning for Multimodal Psychological Counseling}

\author{
  {\large\bfseries
  Jiangnan Yang$^{1,3,\dagger}$~~
  Junjie Chen$^{2,3,\dagger}$~~
  Fei Wang$^{2,3,*}$~~
  Yiqi Nie$^{1,3}$~~
  Yuxin Liu$^{1,3}$\\
  \large\bfseries
  Zhangling Duan$^{3,*}$~~
  Jie Chen$^{1}$~~
  } \\ \vspace{0.3em}
  {\normalsize\normalfont
    $^1$Anhui University~~
    $^2$Hefei University of Technology \\
    $^3$Institute of Artificial Intelligence, Hefei Comprehensive National Science Center\\
    \vspace{0.3em}
    $\dagger$Equal contribution. $^*$Corresponding authors.\\
  }
}

\begin{document}

\maketitle

\begin{abstract}
Psychological counseling is a fundamentally multimodal cognitive process in which clinicians integrate verbal content with visual and vocal cues to infer clients' mental states and respond empathically.
However, most existing language-model-based counseling systems operate on text alone and rely on implicit mental state inference.
We introduce DELTA, a deliberative multi-agent framework that models counseling as a structured reasoning process over multimodal signals, separating evidence grounding, mental state abstraction, and response generation.
DELTA further incorporates reinforcement learning guided by a distribution-level Emotion Attunement Score to encourage emotionally attuned responses.
Experiments on a multimodal counseling benchmark show that DELTA improves both counseling quality and emotion attunement across models.
Ablation and qualitative analyses suggest that explicit multimodal reasoning and structured mental state representations play complementary roles in supporting empathic human-AI interaction.

\end{abstract}

\section{Instruction}
\textcolor{red}{\textit{\textbf{
Note. This work explores multimodal and multi-agent approaches to assist psychological counseling, rather than replacing mental health practitioners or clinical care.
}}}

Psychological counseling is inherently multimodal.
Beyond verbal content, counselors routinely attend to visual and vocal cues, such as facial expressions, tone of voice, and affective fluctuations, to infer clients' mental states and respond with appropriate empathy.
These nonverbal signals often convey critical emotional information that is absent or ambiguous in text alone, particularly in emotionally charged or uncertain situations~\citep{westland2015verbal,marcoux2024nonverbal}.

\begin{figure}[t]
\begin{center}
\includegraphics[width=1\linewidth]{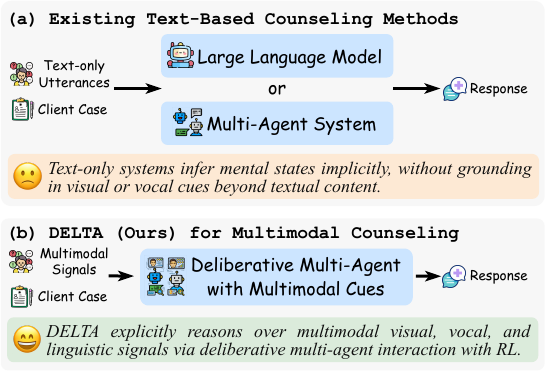}
\caption{
\textbf{Comparison between existing methods and DELTA~(Ours).}
Existing methods adapt large language models or multi-agent systems for psychological counseling primarily by processing client cases and text-only utterances, thereby requiring the inference of mental states and overlooking visual and vocal cues.
In contrast, \textit{DELTA~(Ours) integrates multimodal client signals and supports explicit deliberative multi-agent reasoning with reinforcement learning~(RL),} leading to more grounded counseling responses.
}
\label{fig:motivation}
\end{center}
\end{figure}

Recent advances in large language models (LLMs) have enabled promising progress in automated psychological counseling and emotional support.
Existing approaches typically adapt LLMs or multi-agent systems through prompt engineering, role conditioning, or structured dialogue, demonstrating improvements in fluency and domain relevance~\citep{chu2025towards,chen2023soulchat,lee2024cactus}.
However, as illustrated in Figure~\ref{fig:motivation}, most prior methods primarily operate on client cases and text-only utterances, requiring mental states to be inferred implicitly without explicit grounding in visual or vocal cues~\citep{lee2024cactus,bi2025magi,chen2025mind,dai2025psyche,hu2024psycollm}.
As a result, these systems may produce linguistically appropriate responses that lack emotional grounding or empathic depth.

This gap highlights two key challenges.
First, how to explicitly ground counseling reasoning in multimodal behavioral evidence in a manner that is interpretable and aligned with psychological practice.
Second, how to optimize counseling responses toward emotional attunement, a nuanced objective that cannot be adequately captured by supervised learning alone~\citep{gottman2011science}.
Addressing these challenges requires both a structured reasoning framework and a learning signal that reflects emotionally appropriate counseling behavior.

In this work, we propose \textbf{DELTA}~(\textbf{DEL}iberative mul\textbf{T}i-\textbf{A}gent), a framework that conceptualizes psychological counseling as a structured reasoning process over multimodal client signals.
DELTA is motivated by the observation that effective counseling involves iterative hypothesis formation and verification over heterogeneous cues, rather than a single-pass inference~\citep{haynes2011scientific,norcross2005handbook}.
Accordingly, DELTA explicitly separates multimodal evidence grounding, mental state abstraction, and response generation through deliberative interaction among specialized agents, enabling interpretable and psychologically informed reasoning.
Specifically, DELTA employs dedicated agents to elicit and verify visual and vocal cues from multimodal inputs, consolidates this evidence into a structured mental state representation inspired by clinical mental state examination practices~\citep{Huline-Dickens2013}, and generates counseling responses grounded in this representation.
Finally, to ensure emotionally appropriate response realization, we optimize the response generation policy using reinforcement learning guided by a distribution-level \emph{Emotion Attunement Score}~(EAS).
This design is motivated by psychological theories emphasizing that effective counseling requires \emph{emotion attunement}, namely emotional expressions that are compatible with the client’s current emotional state without rigid mirroring~\citep{greenberg2007emotion,gottman2011science}, a property that is difficult to specify with supervised objectives alone.

We evaluate DELTA on a multimodal emotional support benchmark and demonstrate consistent improvements in both counseling quality and emotion attunement across a range of proprietary and open-source language models.
Importantly, we show that the proposed multi-agent multimodal reasoning workflow alone yields substantial gains over direct prompting, and that reinforcement learning with EAS further strengthens emotional grounding and empathic expression.
Qualitative analyses further illustrate that DELTA produces responses with richer empathic expressions and clearer emotional understanding.
Overall, DELTA bridges psychological theory and modern AI systems by explicitly modeling how multimodal evidence, structured mental state reasoning, and reinforcement learning jointly contribute to effective counseling interactions.
Our main contributions are threefold:
\begin{itemize}
\item We propose DELTA, a deliberative multi-agent framework that explicitly grounds psychological counseling in multimodal visual, vocal, and linguistic evidence, enabling interpretable and psychologically informed reasoning.
\item We introduce the Emotion Attunement Score (EAS), a distribution-level metric for evaluating and optimizing emotional compatibility between counseling responses and clients' multimodal emotional states.
\item We demonstrate that reinforcement learning guided by EAS effectively enhances both emotion attunement and counseling quality, validated through extensive quantitative and qualitative studies.
\end{itemize}

\begin{figure*}[t]
\begin{center}
\includegraphics[width=1\linewidth]{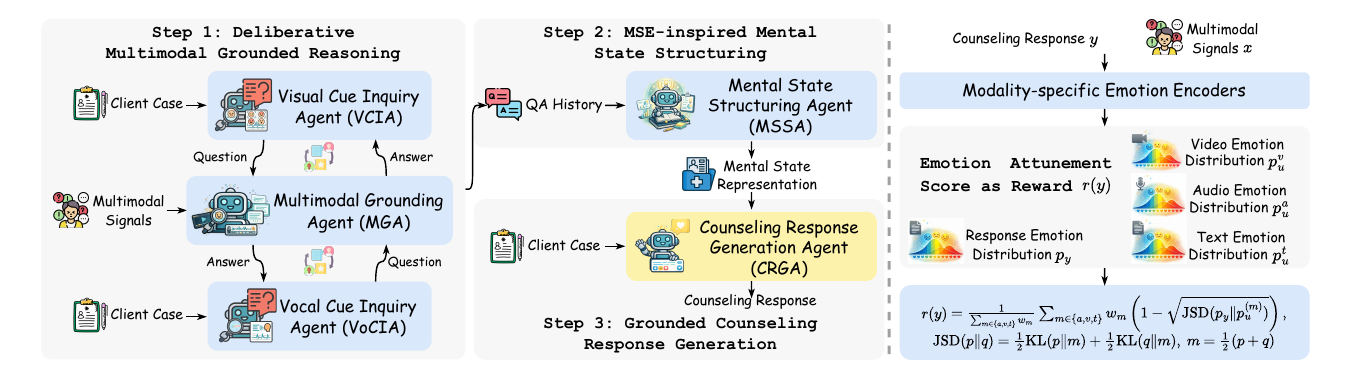}
\caption{
\textbf{DELTA Framework Overview.}
\textbf{\textit{Left:}} Visual and vocal cue inquiry agents pose modality-specific questions, which are answered by a multimodal grounding agent using audio-visual evidence to form a cross-modal QA history.
This history is summarized into a structured mental state representation inspired by the Mental State Examination (MSE) and fed to the counseling response generation agent, optimized via GRPO, a reinforcement learning method, to encourage emotion-attuned responses.
\textbf{\textit{Right:}} Correspondingly, the Emotion Attunement Score (EAS) measures Jensen-Shannon distance between the response emotion distribution and multimodal client emotion distributions, and serves as the reward for GRPO.
}
\label{fig:delta}
\end{center}
\end{figure*}

\section{Related Work}
\paragraph{LLMs for Psychological Counseling.}
Recent progress in psychological counseling with large language models (LLMs) has primarily focused on adapting general-purpose models through instruction tuning, prompting strategies, and domain-specific data.
Representative efforts include long-form counseling corpora such as PsyQA~\citep{sun2021psyqa}, extensions from one-shot assistance to inclusive multi-turn conversations~\citep{qiu2024smile}, and empathy-oriented fine-tuning frameworks such as SoulChat~\citep{chen2023soulchat}, which substantially improve empathic language generation and supportive response quality.
More recently, \emph{theory-informed counseling} and \emph{professional-like reasoning} have gained increasing attention.
CBT-grounded modeling and evaluation align LLM behaviors with therapeutic procedures and strategies, while report-based reconstruction supports more realistic multi-turn assessment~\citep{lee2024cactus,na2024cbt,zhang2024cpsycoun}.
In parallel, diagnostic criteria and therapeutic frameworks have been incorporated into system objectives and designs to improve clinical alignment, interpretability, and safety, often in conjunction with multi-agent deliberation and interactive, feedback-driven optimization~\citep{xu2025autocbt,ozgun2025trustworthy,hu2025agentmental,bi2025magi,chen2025mind,zhu2025psi}.
Despite these advances, most existing systems remain text-only and rely on implicit mental state inference.
Our approach instead explicitly grounds counseling reasoning in multimodal behavioral evidence and maintains structured intermediate representations.

\paragraph{Multimodal Psychological Counseling.}
Multimodal modeling is promising for psychological counseling, as clinicians rely on non-verbal signals such as facial expressions and vocal characteristics.
Existing work mainly falls into two categories: multimodal assessment, including depression detection~\citep{wu2024vs} and explainable emotion or stress reasoning~\citep{wang2024cognition}, and emotional support generation that couples affect recognition with response strategies~\citep{chu2025towards,na-etal-2025-survey}.
However, multimodal counseling remains challenged by noisy cues and limited interpretability.
Our work addresses this by explicitly decomposing multimodal reasoning and maintaining traceable mental state representations rather than opaque feature fusion.

\paragraph{Reinforcement Learning for Alignment.}
Reinforcement learning (RL) and preference-based alignment have been explored to improve counseling quality by combining \emph{accuracy rewards} for clinically grounded content with \emph{format rewards} that encourage professional interaction patterns~\citep{dai2025psyche,guo2025deepseek}.
Theory-grounded supervision further enhances controllability and verification beyond generic preference optimization~\citep{lee2024cactus,iftikhar2024therapy,blackburn2001revised}.
Nevertheless, most existing alignment approaches rely on text-only and label-level objectives.
In contrast, our method optimizes distribution-level emotional attunement grounded in multimodal signals.

\section{Method}

We consider the task of multimodal psychological counseling.
Given a client case context $c$ and multimodal client signals $x$ including visual, vocal, and linguistic information, the goal is to generate a supportive counseling response $y$.

The DELTA framework decomposes counseling into explicit multimodal reasoning, mental state abstraction, and grounded response generation.
An overview of the framework is shown in Figure~\ref{fig:delta}.

\subsection{Agent Roles}

DELTA adopts a functional decomposition of the counseling process by instantiating a set of specialized agents, each responsible for a distinct role.
This design separates evidence grounding, modality-specific inquiry, mental state abstraction, and response generation, enabling grounded and interpretable reasoning over multimodal inputs.
We denote the multimodal client signals as $x=\{x^{(v)}, x^{(a)}, x^{(t)}\}$, corresponding to visual, vocal, and linguistic modalities, respectively.

\paragraph{\raisebox{-.2\baselineskip}{\includegraphics[height=1.3\baselineskip]{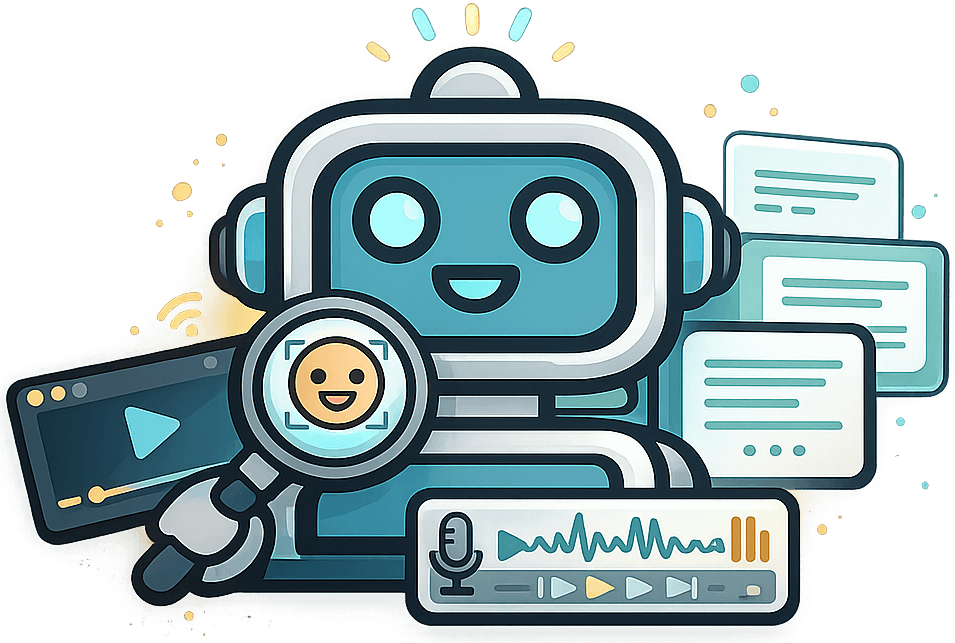}}~Multimodal Grounding Agent (MGA).}
MGA serves as the perceptual backbone of DELTA and is the only component with direct access to audio-visual signals.
Given a query $q$, MGA analyzes $x^{(v)}$ and $x^{(a)}$ to produce an evidence-grounded answer $a$.
By centralizing perceptual access within MGA, DELTA ensures that downstream reasoning remains anchored in observable multimodal evidence.

\paragraph{\raisebox{-.2\baselineskip}{\includegraphics[height=1.3\baselineskip]{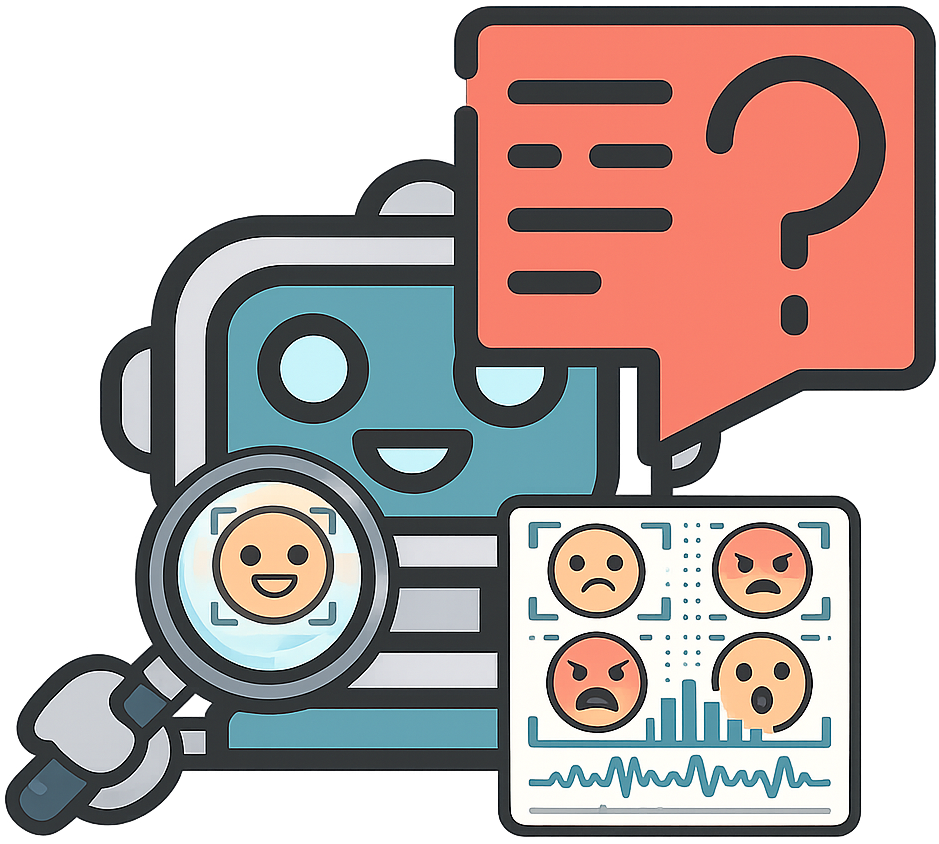}}~Visual Cue Inquiry Agent (VCIA).}
VCIA formulates modality-specific queries $q^{(v)}$ that probe visually grounded information relevant to the client’s current mental state.
Conditioned on the case context $c$ and the accumulated interaction history, VCIA directs attention toward visual evidence that complements linguistic information and supports grounded psychological interpretation.

\paragraph{\raisebox{-.2\baselineskip}{\includegraphics[height=1.3\baselineskip]{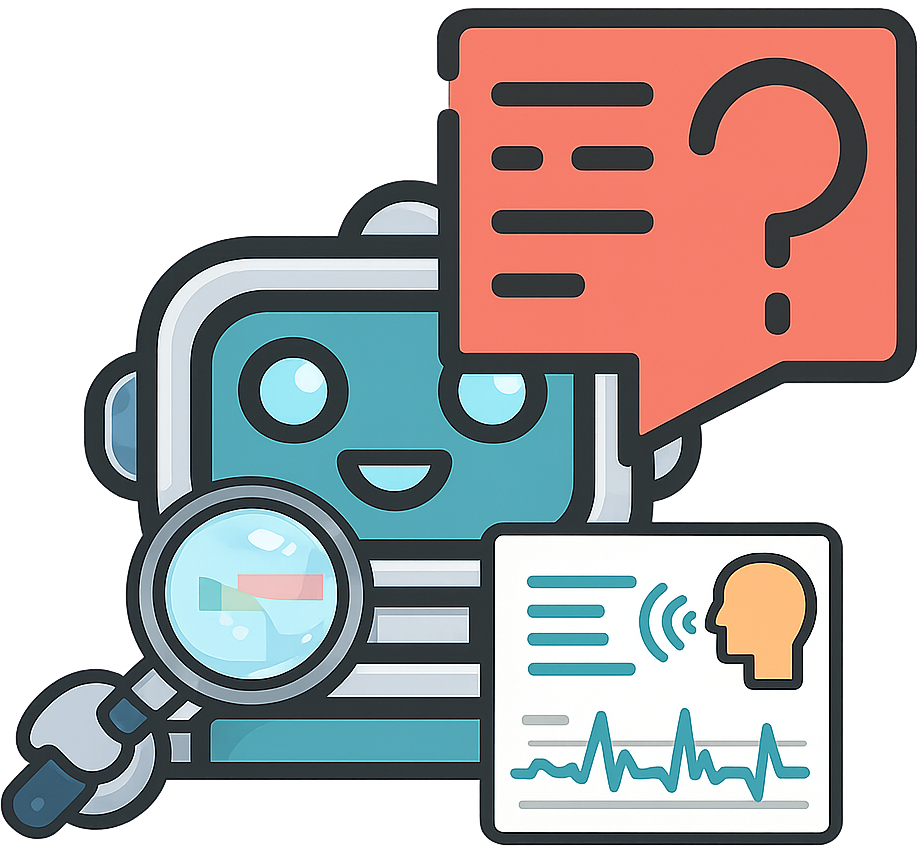}}~Vocal Cue Inquiry Agent (VoCIA).}
VoCIA generates queries $q^{(a)}$ focused on vocal and paralinguistic aspects of the client’s expression.
These cues often convey affective information that is underdetermined by lexical content alone.
By eliciting speech-related evidence, VoCIA enables DELTA to reason about emotional states using information that is complementary to both visual and textual modalities.

\paragraph{\raisebox{-0.2\baselineskip}{\includegraphics[height=1.3\baselineskip]{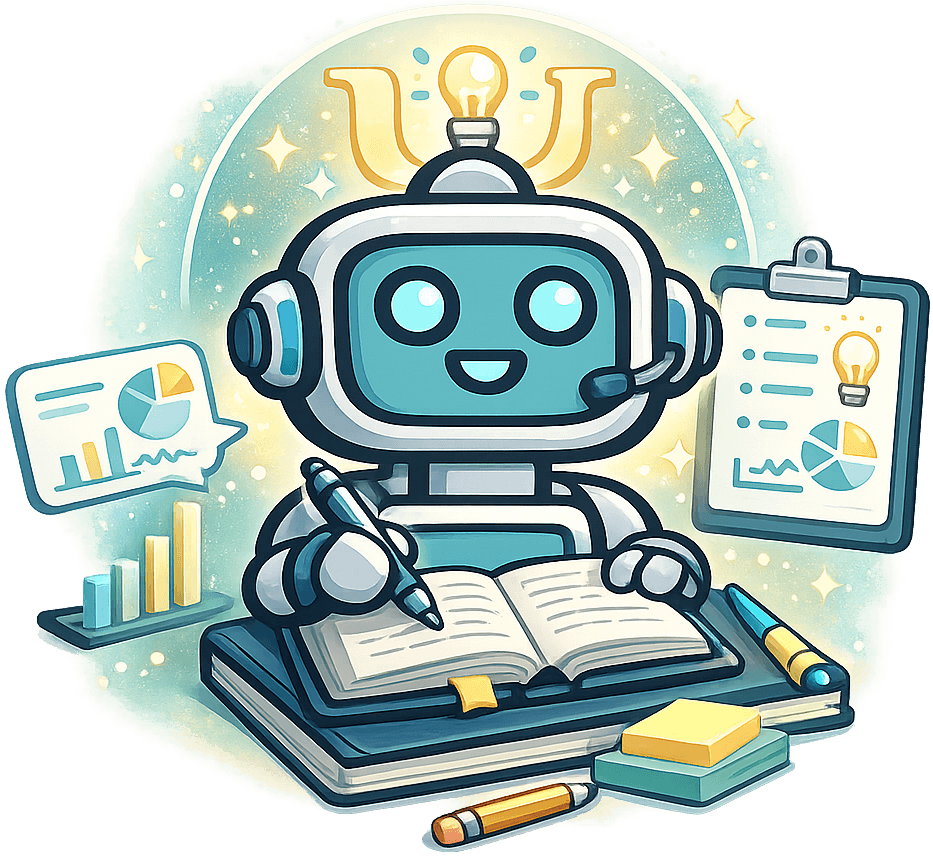}}~Mental State Structuring Agent (MSSA).}
MSSA takes as input the accumulated question-answer history
$H=\{(q^{(v)},a^{(v)}),(q^{(a)},a^{(a)})\}$ and produces a structured mental state representation $M$.
This abstraction step consolidates dispersed multimodal evidence into a coherent intermediate representation inspired by clinical mental-state examination practices \citep{Huline-Dickens2013}, providing a stable interface between evidence elicitation and response generation.

\paragraph{\raisebox{-.1\baselineskip}{\includegraphics[height=1.3\baselineskip]{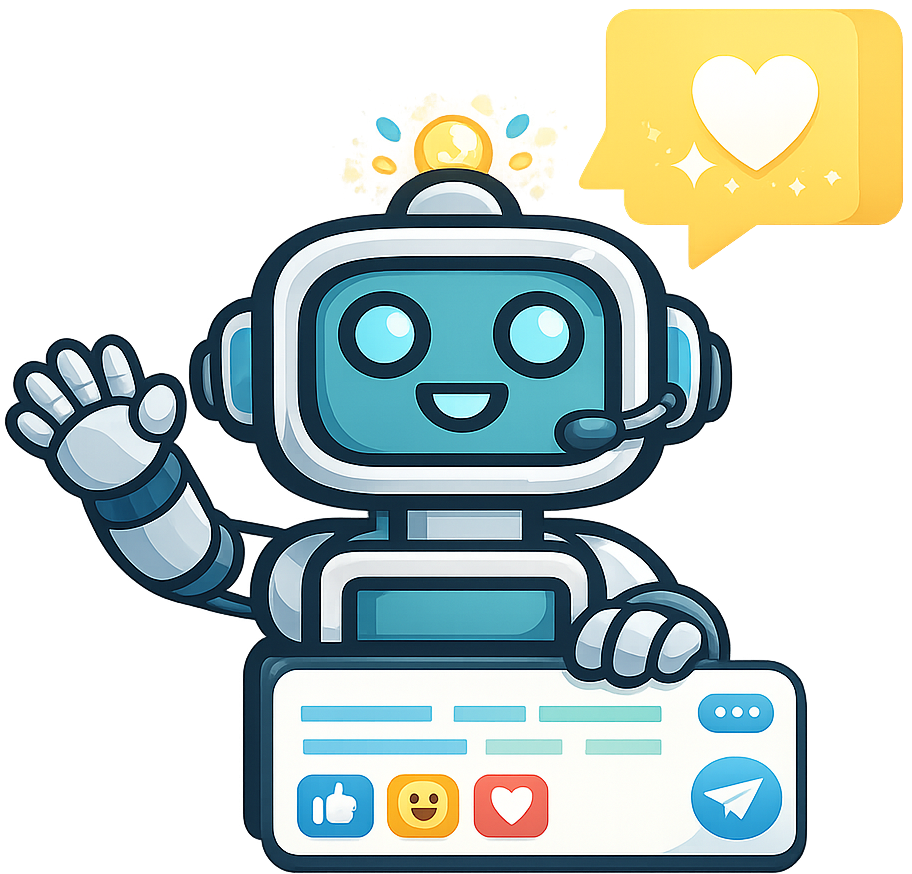}}\hspace{.4\baselineskip}Counseling Response Generation Agent (CRGA).}
CRGA generates the final counseling response $y$ conditioned on the case context $c$ and the structured mental state representation $M$, i.e., $y \sim \pi(\cdot \mid c, M)$.
By operating on abstracted information about mental states rather than raw multimodal inputs, CRGA ensures that response generation is grounded in structured evidence and decoupled from perceptual noise.

\subsection{Workflow}

DELTA operates through a three-stage workflow, illustrated in Figure~\ref{fig:delta}, which organizes multimodal evidence elicitation, mental state abstraction, and response generation into a structured reasoning process.

\paragraph{Step 1: Deliberative Multimodal Grounded Reasoning.}
Given the case context $c$ and multimodal signals $x$, VCIA and VoCIA iteratively generate modality-specific queries $q^{(v)}$ and $q^{(a)}$.
For each query, MGA analyzes the corresponding audio-visual inputs and returns an evidence-grounded answer.
This interaction produces a cross-modal question-answer history
$H=\{(q^{(v)},a^{(v)}),(q^{(a)},a^{(a)})\}$,
which explicitly links mental-state hypotheses to observable multimodal cues.

\paragraph{Step 2: MSE-inspired Mental State Structuring.}
The QA history $H$ is passed to MSSA, which performs mental state abstraction by consolidating multimodal evidence into a structured mental state representation $M$.
This representation summarizes salient affective and behavioral information in a traceable form, serving as an explicit intermediate state for downstream reasoning.

\paragraph{Step 3: Grounded Counseling Response Generation.}
Conditioned on the case context $c$ and the structured mental state representation $M$, CRGA generates the counseling response $y$.
By grounding generation in $M$, the response remains consistent with the elicited multimodal evidence while allowing flexibility in linguistic realization.
The optimization of response behavior is described in the following section.

\subsection{GRPO for Counseling Response Generation}

To encourage more empathic counseling responses, we apply \emph{Group Relative Policy Optimization}~(GRPO)~\citep{shao2024deepseekmath} to fine-tune CRGA.
Reinforcement learning is restricted to CRGA, while all upstream components for evidence grounding and mental state abstraction are kept fixed, as these components encode perceptual and structural priors that should remain stable during response optimization.
 
\paragraph{Policy Formulation.}
CRGA defines a stochastic response policy $\pi_\theta(y \mid c, M)$, where $c$ denotes the client case context and $M$ the structured mental state representation.
Reinforcement learning is applied exclusively to this policy, leaving all upstream grounding and structuring agents fixed.
This design isolates optimization to response realization rather than to perceptual inference or mental-state abstraction.

\paragraph{Emotion Distribution Estimation.}
To estimate emotion distributions for both client signals and generated responses, we employ three modality-specific emotion encoders corresponding to visual, vocal, and linguistic inputs.
Each encoder produces a vector of classification logits over a shared emotion label set $\mathcal{E}$, which is converted into a normalized emotion distribution.
For the generated counseling response $y$, the linguistic emotion encoder yields the response emotion distribution $p_y$.
For the client, modality-specific emotion distributions $\{p_u^{(m)}\}_{m \in \mathcal{M}}$, where $\mathcal{M}=\{v,a,t\}$, are computed independently from visual, vocal, and linguistic signals.
Using full emotion distributions rather than hard labels preserves uncertainty in emotion estimation and provides a smoother signal for downstream optimization.

\paragraph{Emotion Attunement Score~(EAS) as Reward.}
We introduce EAS as a distribution-level metric to evaluate the emotional compatibility between the generated response and the client’s multimodal emotional context, and use it as the reward signal for response optimization.
As shown in the right panel of Figure~\ref{fig:delta}, EAS computes pairwise similarity between the response emotion distribution $p_y$ and each modality-specific client emotion distribution $p_u^{(m)}$ using Jensen-Shannon distance~\citep{menendez1997jensen}, and aggregates these similarities across modalities to form the final reward.

Given two distributions $p$ and $q$, the Jensen-Shannon divergence is defined as:
\begin{equation}
\mathrm{JSD}(p \,\|\, q)
\;=\;
\frac{1}{2}\mathrm{KL}(p \,\|\, m)
+
\frac{1}{2}\mathrm{KL}(q \,\|\, m),
\quad
m = \frac{1}{2}(p+q),
\end{equation}
where $\mathrm{KL}(\cdot\|\cdot)$ denotes the Kullback-Leibler divergence.
The corresponding Jensen-Shannon distance is given by
\begin{equation}
\mathrm{JSDist}(p,q) = \sqrt{\mathrm{JSD}(p \,\|\, q)}.
\end{equation}

We convert distance into a similarity score and aggregate across modalities to form the reward:
\begin{equation}
r(y)
\;=\;
\frac{\sum_{m \in \mathcal{M}} w_m \, \bigl(1 - \mathrm{JSDist}(p_y, p_u^{(m)})\bigr)}
{\sum_{m \in \mathcal{M}} w_m},
\end{equation}
where $w_m \ge 0$ controls the contribution of each modality.

\paragraph{Optimization.}
The response policy is optimized using GRPO by maximizing the expected affective consistency reward:
\begin{equation}
\theta^{*}
\;=\;
\arg\max_{\theta}
\;\mathbb{E}_{y \sim \pi_\theta(\cdot \mid c, M)} \bigl[ r(y) \bigr].
\end{equation}
By optimizing a distribution-level affective objective rather than discrete emotion labels, GRPO encourages empathetic and emotionally attuned counseling responses, rather than rigid emotion matching.

\begin{table*}[t]
\centering
\setlength{\tabcolsep}{2pt}
\begin{minipage}[t]{0.51\linewidth}
\centering
\caption{
\textbf{Comparison of counseling quality with the Direct Prompting (DP) baseline on MESC.}
Scores are reported for Comprehensiveness (Comp.), Professionalism (Prof.), Authenticity (Auth.), Safety (Safe.), and their aggregate (Agg.).
Bold model names indicate DP results, while \texttt{+DELTA} denotes the same base model augmented with our framework.
\textcolor{green!10}{\rule{1.5em}{0.8em}} indicates improvement or no change over DP, and \textcolor{red!10}{\rule{1.5em}{0.8em}} indicates degradation.
}
\label{tab:main-counseling-results}
\vskip 0.12in
\resizebox{\linewidth}{!}{%
\begin{tabular}{
l
>{\columncolor{gray!12}}c
c
>{\columncolor{gray!12}}c
c
>{\columncolor{gray!12}}c
}
\rowcolor{gray!40}
\headcell{Model}{}
& \headcell{Comp.}{}
& \headcell{Prof.}{} 
& \headcell{Auth.}{}
& \headcell{Safe.}{}
& \headcell{Agg.}{} \\
MESC
& 8.71 & 29.10 & 32.50 & 88.56 & 39.72\\

\rowcolor{gray!25}
\multicolumn{6}{c}{\textit{\textbf{Proprietary Models}}}  \\
\textbf{GPT-4o}
& 45.92 & 75.15 & 73.10 & 100.00 & 73.54 \\
\texttt{+DELTA}
& 48.62\updiff{+2.70}
& 79.56\updiff{+4.41}
& 76.92\updiff{+3.82}
& 100.00\updiff{+0.00}
& 76.28\updiff{+2.74} \\

\textbf{GPT-5.2}
& 47.50 & 86.75 & 86.03 & 100.00 & 80.07 \\
\texttt{+DELTA}
& 50.00\updiff{+2.50}
& 92.53\updiff{+5.78}
& 91.11\updiff{+5.08}
& 100.00\updiff{+0.00}
& 83.41\updiff{+3.34} \\

\rowcolor{gray!25}
\multicolumn{6}{c}{\textit{\textbf{Open-source Models}}}  \\
\textbf{Hunyuan~(7B)}   
& 26.70 & 56.07 & 57.28 & 100.00 & 60.01 \\
\texttt{+DELTA}
& 34.95\updiff{+8.25}
& 66.50\updiff{+10.43}
& 64.08\updiff{+6.80}
& 100.00\updiff{+0.00}
& 66.38\updiff{+6.37} \\

\textbf{Mistral~(7B)}   
& 37.38 & 64.08 & 62.78 & 100.00 & 66.06 \\
\texttt{+DELTA}
& 41.75\updiff{+4.37}
& 71.36\updiff{+7.28}
& 66.34\updiff{+3.56}
& 99.03\updiff{-0.97}
& 69.62\updiff{+3.56} \\

\textbf{Llama 3.1~(8B)}  
& 45.57 & 71.12 & 66.99 & 99.03 & 70.68 \\
\texttt{+DELTA}
& 46.12\updiff{+0.55}
& 72.33\updiff{+1.21}
& 68.28\updiff{+1.29}
& 99.03\updiff{+0.00}
& 71.44\updiff{+0.76} \\

\textbf{Qwen3~(8B)}     
& 46.68 & 75.99 & 70.61 & 100.00 & 73.32 \\
\texttt{+DELTA}
& 47.43\updiff{+0.75}
& 80.10\updiff{+4.11}
& 76.05\updiff{+5.44}
& 100.00\updiff{+0.00}
& 75.89\updiff{+2.57}
\end{tabular}
}
\end{minipage}\hfill
\begin{minipage}[t]{0.42\linewidth}
\centering
\caption{
\textbf{Emotion Attunement Score (EAS) comparison with the Direct Prompting (DP) baseline.}
EAS is reported for video, audio, text, and their aggregate (Agg.).
Bold model names correspond to DP, and \texttt{+DELTA} denotes results with our framework.
\textcolor{green!10}{\rule{1.5em}{0.8em}} indicates improvement or no change over DP, and \textcolor{red!10}{\rule{1.5em}{0.8em}} indicates degradation.
}
\label{tab:main-emotion-results}
\vskip 0.12in
\centering
\resizebox{\linewidth}{!}{%
\begin{tabular}{
l
>{\columncolor{gray!12}}c
c
>{\columncolor{gray!12}}c
c
}
\rowcolor{gray!40}
\headcell{Model}{}
& \headcell{Video}{}
& \headcell{Audio}{} 
& \headcell{Text}{}
& \headcell{Agg.}{} \\
MESC
& 26.20 & 30.89 & 55.77 & 37.62\\

\rowcolor{gray!25}
\multicolumn{5}{c}{\textit{\textbf{Proprietary Models}}}  \\
\textbf{GPT-4o}
& 31.87 & 37.73 & 40.96 & 36.85\\
\texttt{+DELTA}
& 29.78\updiff{-2.09} & 37.93\updiff{+0.20} & 43.34\updiff{+2.38} & 37.02\updiff{+0.17}\\

\textbf{GPT-5.2}
& 29.37 & 33.84 & 51.68 & 38.30\\
\texttt{+DELTA}
& 30.58\updiff{+1.21} & 36.52\updiff{+2.68} & 60.11\updiff{+8.43} & 42.40\updiff{+4.10}\\

\rowcolor{gray!25}
\multicolumn{5}{c}{\textit{\textbf{Open-source Models}}}  \\

\textbf{Hunyuan~(7B)}   
& 35.19 & 41.32 & 52.15 & 42.89\\
\texttt{+DELTA}
& 35.27\updiff{+0.08} & 42.16\updiff{+0.84} & 53.89\updiff{+1.74} & 43.77\updiff{+0.88}\\

\textbf{Mistral~(7B)}   
& 40.84 & 45.84 & 45.83 & 44.17\\
\texttt{+DELTA}
& 40.03\updiff{-0.81} & 46.50\updiff{+0.66} & 44.15\updiff{-1.68} & 43.56\updiff{-0.61}\\

\textbf{Llama 3.1~(8B)}  
& 30.45 & 36.67 & 48.25 & 38.46\\
\texttt{+DELTA}
& 31.71\updiff{+1.26} & 37.80\updiff{+1.13} & 48.13\updiff{-0.12} & 39.21\updiff{+0.75}\\

\textbf{Qwen3~(8B)}     
& 31.07 & 38.48 & 49.61 & 39.72\\
\texttt{+DELTA}
& 31.66\updiff{+0.59} & 38.71\updiff{+0.23} & 49.99\updiff{+0.38} & 40.12\updiff{+0.40}
\end{tabular}
}
\end{minipage}
\end{table*}

\section{Experiments}

\subsection{Experimental Setup}

\paragraph{DELTA Default Configuration.}
Unless otherwise specified, DELTA is evaluated under a fixed default configuration.
The agents VCIA, VoCIA, and MSSA are implemented using \texttt{Qwen3-8B}\footnote{\url{https://hf.co/Qwen/Qwen3-8B}}, while MGA is implemented using \texttt{Qwen3-Omni-30B-A3B-Instruct}\footnote{https://hf.co/Qwen/Qwen3-Omni-30B-A3B-Instruct} to provide reliable audio-visual grounding.
These upstream agents are kept fixed across all experiments to avoid confounding effects from variations in perceptual or reasoning capacity.

CRGA is the only component whose underlying model is varied for comparative evaluation, enabling controlled assessment of response generation strategies under identical grounded evidence and mental state representations.
By default, the deliberative reasoning stage runs for two interaction rounds, with each inquiry agent issuing one modality-specific query per round, unless stated otherwise.

\paragraph{Dataset and Evaluation.}
We conduct experiments on the MESC dataset~\citep{chu2025towards}, a multimodal emotional support conversation benchmark derived from \textit{In Treatment}, which provides synchronized video, audio, and text data with utterance-level annotations of client emotions and counselor strategies.
We evaluate DELTA from two aspects: counseling quality and emotion attunement.
For counseling quality, we follow the benchmark protocol of~\citeauthor{hu2025beyond}~\citep{hu2025beyond}, which assesses single-turn counseling responses along four dimensions: \textit{Comprehensiveness} (Comp.), \textit{Professionalism} (Prof.), \textit{Authenticity} (Auth.), and \textit{Safety} (Safe.).
These quality scores are obtained using \textit{Gemini-2.0-Flash}\footnote{\url{https://blog.google/innovation-and-ai/models-and-research/google-deepmind/google-gemini-ai-update-december-2024}} as the evaluation model.
Emotion attunement is measured by the proposed \textit{Emotion Attunement Score (EAS)}, computed directly from modality-specific emotion encoders over video, audio, and text.
All reported scores are linearly normalized to $[0,100]$, where higher values indicate better performance.

\paragraph{CRGA Training Details.}
CRGA is optimized on the training split of the MESC dataset using GRPO.
Emotion distributions used for training are estimated with modality-specific pretrained emotion encoders for text, speech, and vision, including a RoBERTa-GoEmotions classifier\footnote{\url{https://hf.co/SamLowe/roberta-base-go_emotions}}, a wav2vec2-based speech emotion recognition model\footnote{\url{http://doi.org/10.57967/hf/2045}}, and a facial-expression recognition model\footnote{\url{https://hf.co/HardlyHumans/Facial-expression-detection}}.
All encoder outputs are projected into a shared 8-way emotion space to enable cross-modal comparison, with predefined label mapping and probability aggregation applied to calibrate encoders whose original label spaces do not align with this target space.
For each training instance, we sample a group of $G=4$ candidate responses and update the policy with a learning rate of $1\times10^{-5}$.
To reduce optimization cost, we adopt LoRA-based parameter-efficient fine-tuning with rank $r=64$, scaling factor $\alpha=128$, and dropout rate $0.05$.
For the Emotion Attunement Score, audio, visual, and textual components are equally weighted, i.e., $w_a = w_v = w_t = \tfrac{1}{3}$, unless stated otherwise.

\paragraph{Additional Experimental Details.}
Further implementation details are provided in the anonymous supplementary material: \url{https://osf.io/ke9yf/overview?view_only=e3d99cb3d26e4326970b167303a4f9cf}.
This includes prompt templates for all agent roles, structured input and output field definitions, and preprocessing details necessary for full reproducibility.

\subsection{Experimental Results}

\begin{figure}[t]
\begin{center}
\includegraphics[width=1\linewidth]{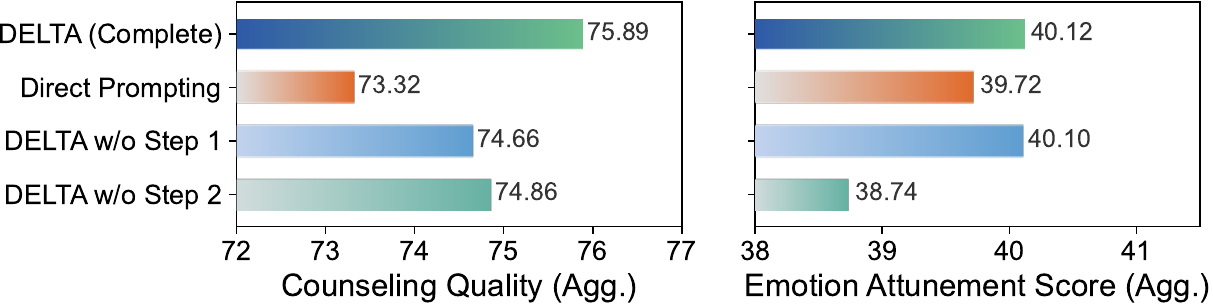}
\caption{
\textbf{Ablation study on DELTA.}
Removing either deliberative multimodal reasoning or mental state structuring leads to consistent performance degradation, while the complete DELTA workflow achieves the best overall results.
}
\label{fig:ablation}
\end{center}
\end{figure}

\paragraph{DELTA Consistently Improves Counseling Quality and Emotion Attunement over Direct Prompting.}
We evaluate the effectiveness of the proposed DELTA workflow by instantiating CRGA with two proprietary models (GPT-4o\footnote{\url{https://openai.com/index/hello-gpt-4o}} and GPT-5.2\footnote{\url{https://openai.com/index/introducing-gpt-5-2}}) and four widely used open-source models (Hunyuan~7B\footnote{\url{https://hf.co/tencent/Hunyuan-7B-Instruct}}, Mistral~7B\footnote{\url{https://hf.co/mistralai/Mistral-7B-Instruct-v0.3}}, Llama~3.1~8B\footnote{\url{https://hf.co/meta-llama/Llama-3.1-8B}}, and Qwen3~8B\footnotemark[1]),
and compare against both the original MESC method~\citep{chu2025towards} and a Direct Prompting (DP) baseline that uses single-prompt role-based generation.
The goal of this experiment is to isolate and assess the contribution of our multi-agent, multimodal reasoning workflow, without applying reinforcement learning to CRGA.
As shown in Table~\ref{tab:main-counseling-results}, DELTA consistently improves counseling quality over DP across most models and evaluation dimensions, with particularly notable gains in Comprehensiveness, Professionalism, and Authenticity, while maintaining high Safety scores.
Table~\ref{tab:main-emotion-results} further shows that DELTA generally improves Emotion Attunement Score (EAS) across video, audio, and text modalities, leading to higher aggregated emotion attunement in most settings.
Overall, these results demonstrate that the DELTA workflow alone, even without reinforcement learning, can yield consistent and meaningful improvements over direct prompting, validating the effectiveness of structured multimodal evidence elicitation and deliberative multi-agent reasoning.

\paragraph{Ablation Analysis.}
We conduct an ablation study using Qwen3~(8B) as CRGA to examine the contribution of individual components in the DELTA workflow.
Figure~\ref{fig:ablation} reports aggregated counseling quality and aggregated Emotion Attunement Score (EAS) under four settings: the complete DELTA framework, Direct Prompting (DP), DELTA without Step~1 (deliberative multimodal grounded reasoning), and DELTA without Step~2 (mental state structuring).
This ablation is designed to isolate the effect of structured multimodal reasoning and mental state abstraction.
Removing either Step~1 or Step~2 leads to consistent degradation in at least one evaluation metric, while the complete DELTA framework achieves the best overall performance.
These results demonstrate that both deliberative multimodal reasoning and mental state structuring are necessary and complementary for improving counseling quality and emotion attunement.

\begin{figure}[t]
\begin{center}
\includegraphics[width=0.76\linewidth]{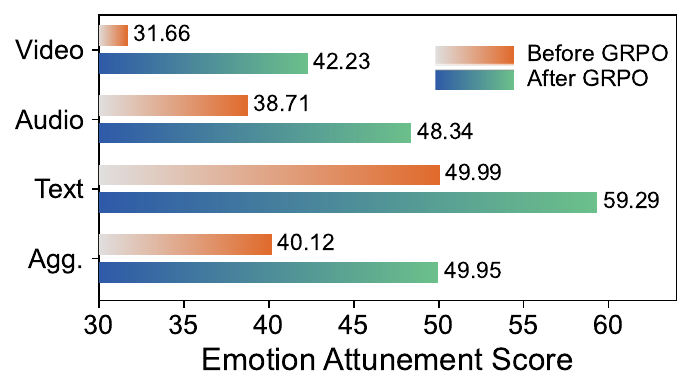}
\caption{
\textbf{Effect of GRPO on EAS.}
GRPO consistently improves emotion attunement across all modalities.
}
\label{fig:grpo}
\end{center}
\end{figure}

\begin{table}[t]
\setlength{\tabcolsep}{2pt}
\caption{
\textbf{Effect of GRPO on counseling quality.}
GRPO leads to consistent improvements in overall counseling quality.
}
\label{tab:grpo}
\vskip 0.12in
\centering
\resizebox{\linewidth}{!}{%
\begin{tabular}{
l
>{\columncolor{gray!12}}c
c
>{\columncolor{gray!12}}c
c
>{\columncolor{gray!12}}c
}
\rowcolor{gray!40}
\headcell{Model}{}
& \headcell{Comp.}{}
& \headcell{Prof.}{} 
& \headcell{Auth.}{}
& \headcell{Safe.}{}
& \headcell{Agg.}{} \\
Before GRPO
& 47.43 & 80.10 & 76.05 & 100.00 & 75.89 \\
After GRPO
& 48.48\updiff{+1.05} & 82.28\updiff{+2.18} & 78.48\updiff{2.43} & 100.00\updiff{+0.00} & 77.31\updiff{+1.42} \\
\end{tabular}
}
\end{table}
\paragraph{GRPO and Emotion Attunement Score Are Effective.}
We further evaluate the effectiveness of GRPO and the proposed Emotion Attunement Score (EAS) by training CRGA with Qwen3~(8B) using GRPO, while keeping the rest of the DELTA workflow unchanged.
Results in Tables~\ref{tab:main-counseling-results} and~\ref{tab:main-emotion-results} reveal a clear correspondence between EAS and counseling quality across models: for instance, Mistral~(7B) exhibits a degradation in EAS when augmented with DELTA, which is accompanied by a consistent drop in aggregated counseling quality, whereas models achieving higher EAS generally obtain improved counseling quality scores.
This alignment suggests that EAS captures emotionally relevant properties that are predictive of counseling effectiveness.
After applying GRPO, Figure~\ref{fig:grpo} shows that EAS improves substantially across all modalities, and Table~\ref{tab:grpo} further demonstrates corresponding gains in counseling quality.
Together, these results indicate that optimizing responses with GRPO under the guidance of EAS effectively enhances both emotion attunement and counseling quality, validating the utility of EAS as a training signal and the effectiveness of GRPO for response optimization.

\begin{figure}[t]
\begin{center}
\includegraphics[width=\linewidth]{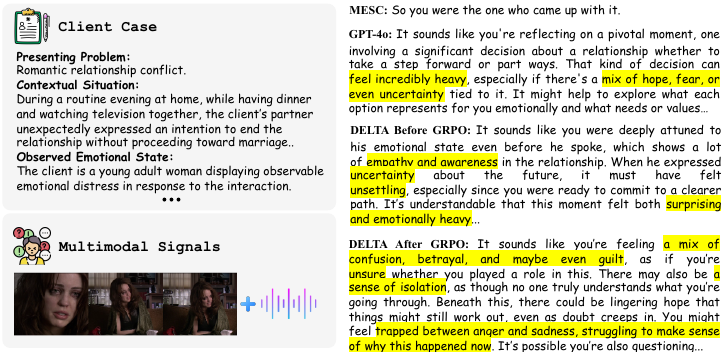}
\caption{
\textbf{Qualitative comparison.}
Under the same client case and multimodal inputs, DELTA produces counseling responses with more explicit empathic expressions and stronger emotional grounding than the baselines, and this advantage is further strengthened after GRPO.
}
\label{fig:case}
\end{center}
\end{figure}

\paragraph{Qualitative Comparison.}
We further conduct a qualitative comparison to examine the behavioral differences between DELTA and baseline methods.
As shown in Figure~\ref{fig:case}, under the same client case and multimodal inputs, DELTA produces counseling responses with more explicit empathic expressions and clearer emotional grounding.
This advantage is already observable before reinforcement learning and is further strengthened after GRPO, complementing the quantitative improvements reported in previous sections.

\section{Conclusion}
We presented DELTA, a deliberative multi-agent framework for multimodal psychological counseling that explicitly grounds reasoning in visual, vocal, and linguistic cues.
By separating evidence grounding, mental state abstraction, and response generation, DELTA provides a structured and interpretable account of how multimodal information can support empathic counseling behavior.
Empirical results across multiple models show that DELTA consistently improves both counseling quality and emotion attunement, with ablation and qualitative analyses highlighting the complementary roles of structured multimodal reasoning and learning-based optimization.
These findings suggest that deliberative multimodal reasoning offers a principled alternative to single-pass, text-only counseling approaches.

Future work will explore richer and more dynamic mental state representations and longer-term interactive settings to better capture the evolving dynamics of real-world counseling.

\printbibliography

@article{xu2025autocbt,
  title={Autocbt: An autonomous multi-agent framework for cognitive behavioral therapy in psychological counseling},
  author={Xu, Ancheng and Yang, Di and Li, Renhao and Zhu, Jingwei and Tan, Minghuan and Yang, Min and Qiu, Wanxin and Ma, Mingchen and Wu, Haihong and Li, Bingyu and others},
  journal={arXiv preprint arXiv:2501.09426},
  year={2025}
}

@article{zhu2025psi,
  title={$\{$$\backslash$Psi$\}$-Arena: Interactive Assessment and Optimization of LLM-based Psychological Counselors with Tripartite Feedback},
  author={Zhu, Shijing and Chen, Zhuang and Bi, Guanqun and Li, Binghang and Deng, Yaxi and Wan, Dazhen and Peng, Libiao and Xiao, Xiyao and Zhang, Rongsheng and Lv, Tangjie and others},
  journal={arXiv preprint arXiv:2505.03293},
  year={2025}
}

@article{hu2025agentmental,
  title={AgentMental: An Interactive Multi-Agent Framework for Explainable and Adaptive Mental Health Assessment},
  author={Hu, Jinpeng and Wang, Ao and Xie, Qianqian and Ma, Hui and Li, Zhuo and Guo, Dan},
  journal={arXiv preprint arXiv:2508.11567},
  year={2025}
}

@article{blackburn2001revised,
  title={The revised cognitive therapy scale (CTS-R): psychometric properties},
  author={Blackburn, Ivy-Marie and James, Ian A and Milne, Derek L and Baker, Chris and Standart, Sally and Garland, Anne and Reichelt, F Katharina},
  journal={Behavioural and cognitive psychotherapy},
  volume={29},
  number={4},
  pages={431--446},
  year={2001}
}

@article{hu2024psycollm,
  title={Psycollm: Enhancing llm for psychological understanding and evaluation},
  author={Hu, Jinpeng and Dong, Tengteng and Gang, Luo and Ma, Hui and Zou, Peng and Sun, Xiao and Guo, Dan and Yang, Xun and Wang, Meng},
  journal={IEEE Transactions on Computational Social Systems},
  year={2024}
}

@inproceedings{wu2024vs,
  title={VS-LLM: Visual-Semantic Depression Assessment Based on LLM for Drawing Projection Test},
  author={Wu, Meiqi and Kang, Yaxuan and Li, Xuchen and Hu, Shiyu and Chen, Xiaotang and Kang, Yunfeng and Wang, Weiqiang and Huang, Kaiqi},
  booktitle={Chinese Conference on Pattern Recognition and Computer Vision (PRCV)},
  pages={232--246},
  year={2024},
  organization={Springer}
}

@article{dai2025psyche,
  title={Psyche-R1: Towards Reliable Psychological LLMs through Unified Empathy, Expertise, and Reasoning},
  author={Dai, Chongyuan and Hu, Jinpeng and Shi, Hongchang and Li, Zhuo and Yang, Xun and Wang, Meng},
  journal={arXiv preprint arXiv:2508.10848},
  year={2025}
}

@inproceedings{bi2025magi,
  title={MAGI: Multi-agent guided interview for psychiatric assessment},
  author={Bi, Guanqun and Chen, Zhuang and Liu, Zhoufu and Wang, Hongkai and Xiao, Xiyao and Xie, Yuqiang and Zhang, Wen and Huang, Yongkang and Chen, Yuxuan and Peng, Libiao and others},
  booktitle={Findings of the Association for Computational Linguistics: ACL 2025},
  pages={24898--24921},
  year={2025}
}

@article{iftikhar2024therapy,
  title={Therapy as an NLP task: psychologists' comparison of LLMs and human peers in CBT},
  author={Iftikhar, Zainab and Ransom, Sean and Xiao, Amy and Nugent, Nicole and Huang, Jeff},
  journal={arXiv preprint arXiv:2409.02244},
  year={2024}
}

@inproceedings{na-etal-2025-survey,
    title = "A Survey of Large Language Models in Psychotherapy: Current Landscape and Future Directions",
    author = "Na, Hongbin  and
      Hua, Yining  and
      Wang, Zimu  and
      Shen, Tao  and
      Yu, Beibei  and
      Wang, Lilin  and
      Wang, Wei  and
      Torous, John  and
      Chen, Ling",
    editor = "Che, Wanxiang  and
      Nabende, Joyce  and
      Shutova, Ekaterina  and
      Pilehvar, Mohammad Taher",
    booktitle = "Findings of the Association for Computational Linguistics: ACL 2025",
    month = jul,
    year = "2025",
    address = "Vienna, Austria",
    publisher = "Association for Computational Linguistics",
    url = "https://aclanthology.org/2025.findings-acl.385/",
    doi = "10.18653/v1/2025.findings-acl.385",
    pages = "7362--7376",
    ISBN = "979-8-89176-256-5"
}

@article{wang2024cognition,
  title={Cognition Chain for Explainable Psychological Stress Detection on Social Media},
  author={Wang, Xin and Gao, Boyan and Dai, Yi and Cao, Lei and Zhao, Liang and Yang, Yibo and Clifton, David},
  journal={arXiv preprint arXiv:2412.14009},
  year={2024}
}

@article{Huline-Dickens2013,
  title   = {The mental state examination},
  author  = {Huline-Dickens, Sarah},
  journal = {Advances in Psychiatric Treatment},
  year    = {2013},
  volume  = {19},
  number  = {2},
  pages   = {97--98},
  doi     = {10.1192/apt.bp.112.010215}
}

@article{guo2025deepseek,
  title={Deepseek-r1: Incentivizing reasoning capability in llms via reinforcement learning},
  author={Guo, Daya and Yang, Dejian and Zhang, Haowei and Song, Junxiao and Zhang, Ruoyu and Xu, Runxin and Zhu, Qihao and Ma, Shirong and Wang, Peiyi and Bi, Xiao and others},
  journal={arXiv preprint arXiv:2501.12948},
  year={2025}
}

@article{chu2025towards,
  title={Towards multimodal emotional support conversation systems},
  author={Chu, Yuqi and Liao, Lizi and Zhou, Zhiyuan and Ngo, Chong-Wah and Hong, Richang},
  journal={IEEE Transactions on Multimedia},
  year={2025},
  publisher={IEEE}
}

@article{hu2025beyond,
  title={Beyond empathy: Integrating diagnostic and therapeutic reasoning with large language models for mental health counseling},
  author={Hu, He and Zhou, Yucheng and Si, Juzheng and Wang, Qianning and Zhang, Hengheng and Ren, Fuji and Ma, Fei and Cui, Laizhong and Tian, Qi},
  journal={arXiv preprint arXiv:2505.15715},
  year={2025}
}

@article{shao2024deepseekmath,
  title={Deepseekmath: Pushing the limits of mathematical reasoning in open language models},
  author={Shao, Zhihong and Wang, Peiyi and Zhu, Qihao and Xu, Runxin and Song, Junxiao and Bi, Xiao and Zhang, Haowei and Zhang, Mingchuan and Li, YK and Wu, Yang and others},
  journal={arXiv preprint arXiv:2402.03300},
  year={2024}
}

@inproceedings{qiu2024smile,
  title={Smile: Single-turn to multi-turn inclusive language expansion via chatgpt for mental health support},
  author={Qiu, Huachuan and He, Hongliang and Zhang, Shuai and Li, Anqi and Lan, Zhenzhong},
  booktitle={Findings of the Association for Computational Linguistics: EMNLP 2024},
  pages={615--636},
  year={2024}
}

@inproceedings{chen2023soulchat,
    title = "{S}oul{C}hat: Improving {LLM}s' Empathy, Listening, and Comfort Abilities through Fine-tuning with Multi-turn Empathy Conversations",
    author = "Chen, Yirong  and
      Xing, Xiaofen  and
      Lin, Jingkai  and
      Zheng, Huimin  and
      Wang, Zhenyu  and
      Liu, Qi  and
      Xu, Xiangmin",
    editor = "Bouamor, Houda  and
      Pino, Juan  and
      Bali, Kalika",
    booktitle = "Findings of the Association for Computational Linguistics: EMNLP 2023",
    month = dec,
    year = "2023",
    address = "Singapore",
    publisher = "Association for Computational Linguistics",
    url = "https://aclanthology.org/2023.findings-emnlp.83/",
    doi = "10.18653/v1/2023.findings-emnlp.83",
    pages = "1170--1183"
}

@inproceedings{chen2025mind,
    title = "{MIND}: Towards Immersive Psychological Healing with Multi-Agent Inner Dialogue",
    author = "Chen, Yujia  and
      Li, Changsong  and
      Wang, Yiming  and
      Ju, Tianjie  and
      Xiao, Qingqing  and
      Zhang, Nan  and
      Kong, Zifan  and
      Wang, Peng  and
      Yan, Binyu",
    editor = "Christodoulopoulos, Christos  and
      Chakraborty, Tanmoy  and
      Rose, Carolyn  and
      Peng, Violet",
    booktitle = "Findings of the Association for Computational Linguistics: EMNLP 2025",
    month = nov,
    year = "2025",
    address = "Suzhou, China",
    publisher = "Association for Computational Linguistics",
    url = "https://aclanthology.org/2025.findings-emnlp.499/",
    doi = "10.18653/v1/2025.findings-emnlp.499",
    pages = "9380--9413",
    ISBN = "979-8-89176-335-7"
}

@inproceedings{lee2024cactus,
  title={Cactus: Towards psychological counseling conversations using cognitive behavioral theory},
  author={Lee, Suyeon and Mac Kim, Sunghwan and Kim, Minju and Kang, Dongjin and Yang, Dongil and Kim, Harim and Kang, Minseok and Jung, Dayi and Kim, Min Hee and Lee, Seungbeen and others},
  booktitle={Findings of the Association for Computational Linguistics: EMNLP 2024},
  pages={14245--14274},
  year={2024}
}

@inproceedings{na2024cbt,
    title = "{CBT}-{LLM}: A {C}hinese Large Language Model for Cognitive Behavioral Therapy-based Mental Health Question Answering",
    author = "Na, Hongbin",
    editor = "Calzolari, Nicoletta  and
      Kan, Min-Yen  and
      Hoste, Veronique  and
      Lenci, Alessandro  and
      Sakti, Sakriani  and
      Xue, Nianwen",
    booktitle = "Proceedings of the 2024 Joint International Conference on Computational Linguistics, Language Resources and Evaluation (LREC-COLING 2024)",
    month = may,
    year = "2024",
    address = "Torino, Italia",
    publisher = "ELRA and ICCL",
    url = "https://aclanthology.org/2024.lrec-main.261/",
    pages = "2930--2940"
}

@inproceedings{ozgun2025trustworthy,
  title={Trustworthy AI Psychotherapy: Multi-Agent LLM Workflow for Counseling and Explainable Mental Disorder Diagnosis},
  author={Ozgun, Mithat Can and Pei, Jiahuan and Hindriks, Koen and Donatelli, Lucia and Liu, Qingzhi and Wang, Junxiao},
  booktitle={Proceedings of the 34th ACM International Conference on Information and Knowledge Management},
  pages={2263--2272},
  year={2025}
}

@inproceedings{sun2021psyqa,
    title = "{P}sy{QA}: A {C}hinese Dataset for Generating Long Counseling Text for Mental Health Support",
    author = "Sun, Hao  and
      Lin, Zhenru  and
      Zheng, Chujie  and
      Liu, Siyang  and
      Huang, Minlie",
    editor = "Zong, Chengqing  and
      Xia, Fei  and
      Li, Wenjie  and
      Navigli, Roberto",
    booktitle = "Findings of the Association for Computational Linguistics: ACL-IJCNLP 2021",
    month = aug,
    year = "2021",
    address = "Online",
    publisher = "Association for Computational Linguistics",
    url = "https://aclanthology.org/2021.findings-acl.130/",
    doi = "10.18653/v1/2021.findings-acl.130",
    pages = "1489--1503"
}

@inproceedings{zhang2024cpsycoun,
    title = "{CP}sy{C}oun: A Report-based Multi-turn Dialogue Reconstruction and Evaluation Framework for {C}hinese Psychological Counseling",
    author = "Zhang, Chenhao  and
      Li, Renhao  and
      Tan, Minghuan  and
      Yang, Min  and
      Zhu, Jingwei  and
      Yang, Di  and
      Zhao, Jiahao  and
      Ye, Guancheng  and
      Li, Chengming  and
      Hu, Xiping",
    editor = "Ku, Lun-Wei  and
      Martins, Andre  and
      Srikumar, Vivek",
    booktitle = "Findings of the Association for Computational Linguistics: ACL 2024",
    month = aug,
    year = "2024",
    address = "Bangkok, Thailand",
    publisher = "Association for Computational Linguistics",
    url = "https://aclanthology.org/2024.findings-acl.830/",
    doi = "10.18653/v1/2024.findings-acl.830",
    pages = "13947--13966"
}

@article{marcoux2024nonverbal,
  title={Nonverbal behaviors perceived as most empathic in a simulated medical context},
  author={Marcoux, Audrey and Tessier, Marie-H{\'e}l{\`e}ne and Jackson, Philip L},
  journal={Computers in Human Behavior},
  volume={157},
  pages={108268},
  year={2024},
  publisher={Elsevier}
}

@book{westland2015verbal,
  title={Verbal and non-verbal communication in psychotherapy},
  author={Westland, Gill},
  year={2015},
  publisher={WW Norton \& Company}
}

@book{gottman2011science,
  title={The science of trust: Emotional attunement for couples},
  author={Gottman, John M},
  year={2011},
  publisher={WW Norton \& Company}
}

@article{greenberg2007emotion,
  title={Emotion in the therapeutic relationship in emotion-focused therapy},
  author={Greenberg, Leslie S},
  journal={The therapeutic relationship in the cognitive behavioral psychotherapies},
  pages={43--62},
  year={2007},
  publisher={Routledge}
}

@article{menendez1997jensen,
  title={The jensen-shannon divergence},
  author={Men{\'e}ndez, Mar{\'\i}a Luisa and Pardo, Julio Angel and Pardo, Leandro and Pardo, Mar{\'\i}a del C},
  journal={Journal of the Franklin Institute},
  volume={334},
  number={2},
  pages={307--318},
  year={1997},
  publisher={Elsevier}
}

@book{haynes2011scientific,
  title={Scientific foundations of clinical assessment},
  author={Haynes, Stephen N and Smith, Gregory T and Hunsley, John D},
  year={2011},
  publisher={Routledge}
}

@book{norcross2005handbook,
  title={Handbook of psychotherapy integration},
  author={Norcross, John C and Goldfried, Marvin R},
  year={2005},
  publisher={Oxford University Press}
}

\end{document}